\documentclass[conference]{IEEEtran}

\ifCLASSINFOpdf
\else
\fi

\usepackage{graphicx}
\usepackage{subfig}
\usepackage[pagebackref=true,breaklinks=true,colorlinks,bookmarks=false]{hyperref}
\usepackage{booktabs}

\usepackage{fancyhdr}

\renewcommand\cap[3]{\caption[#2]{\label{#1}\textsc{#2}. \small\textit{#3}}}

\graphicspath{{./}{./misc/}}

\begin{document}

\title{Are Accuracy and Robustness Correlated?}


\author{\IEEEauthorblockN{Andras Rozsa, Manuel G\"unther, and Terrance E. Boult }
\IEEEauthorblockA{University of Colorado at Colorado Springs\\
Vision and Security Technology (VAST) Lab\\
Email: http://vast.uccs.edu/contact-us
}}



\maketitle

{
  \chead{\small This is a pre-print of the original paper accepted for oral presentation at ICMLA 2016.}
  \thispagestyle{fancy}
}

\begin{abstract}
Machine learning models are vulnerable to adversarial examples formed by applying small carefully chosen perturbations to inputs that cause unexpected classification errors. In this paper, we perform experiments on various adversarial example generation approaches with multiple deep convolutional neural networks including Residual Networks, the best performing models on ImageNet Large-Scale Visual Recognition Challenge 2015. We compare the adversarial example generation techniques with respect to the quality of the produced images, and measure the robustness of the tested machine learning models to adversarial examples. Finally, we conduct large-scale experiments on cross-model adversarial portability. We find that adversarial examples are mostly transferable across similar network topologies, and we demonstrate that better machine learning models are less vulnerable to adversarial examples.
\end{abstract}


\hfill

\section{Introduction}

In the last few years, deep neural networks have become the most powerful machine learning models and have been successfully applied to various tasks. Due to the significant performance improvements on visual recognition tasks, state-of-the-art machine learning models have managed to obtain classification accuracies comparable to or even better than human-level performance \cite{szegedy2015going,he2015delving,chen2016unconstrained,he2015deep}. This performance boost is mainly the result of advances in two technical directions, namely, building more powerful learning models and designing better strategies to avoid overfitting.

Although deep neural networks provide outstanding performance on various recognition tasks, an intriguing property of these models was revealed by Szegedy et al.~\cite{szegedy2013intriguing}. Machine learning models, including state-of-the-art deep neural networks, unexpectedly and confidently misclassify inputs crafted by adding \emph{imperceptible}, non-random perturbations to correctly classified images. These perturbed samples that cause classification errors are called adversarial examples and their existence reveals at least two problems. First, it demonstrates that there is a classification inconsistency between vulnerable machine learning models and human observers, as adversarial images are generally indistinguishable from their corresponding original inputs by human perception. Second, adversarial examples demonstrate that deep neural networks do not generalize well, in other words, they are not robust to small perturbations to their inputs.


\begin{figure}[!t]
\begin{center}
\vspace*{-0.1in}
\centering\subfloat[][\label{fig:adv:a}\centering VGG-16: \textit{white shark} \par \textit{PASS}=$0.9999$, $L_2$=$15.84$, $L_\infty$=$1$]{\includegraphics[width=.49\columnwidth]{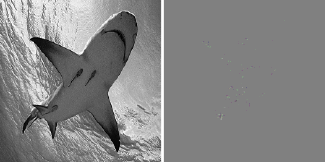}}
\hfill
\centering\subfloat[][\label{fig:adv:b}\centering BVLC-GoogLeNet: \textit{hammerhead} \par \textit{PASS}=$0.9998$, $L_2$=$60.93$, $L_\infty$=$3$]{\includegraphics[width=.49\columnwidth]{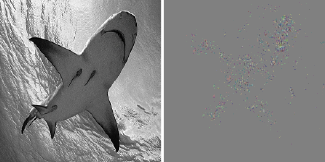}}
\\
\centering\subfloat[][\label{fig:adv:c}\centering ResNet-50: \textit{hammerhead} \par \textit{PASS}=$0.9992$, $L_2$=$111.43$, $L_\infty$=$5$]{\includegraphics[width=.49\columnwidth]{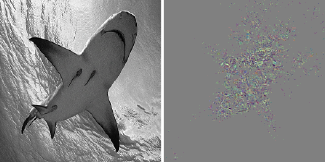}}
\hfill
\centering\subfloat[][\label{fig:adv:d}\centering ResNet-152: \textit{white shark} \par \textit{PASS}=$0.9985$, $L_2$=$165.93$, $L_\infty$=$9$]{\includegraphics[width=.49\columnwidth]{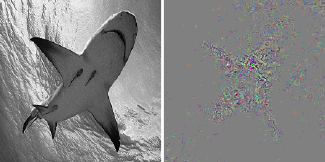}}

\cap{fig:adv}{Adversarial Examples and Perturbations}{This figure shows adversarial examples and their corresponding perturbations for a tiger shark image generated by the fast gradient value (FGV) approach \cite{rozsa2016adversarial} on four deep convolutional neural networks: \subref{fig:adv:a} VGG-16 \cite{Simonyan15}, \subref{fig:adv:b} Berkeley-trained version of GoogLeNet \cite{szegedy2015going} (BVLC-GoogLeNet), \subref{fig:adv:c} Residual Network \cite{he2015deep} with 50 layers (ResNet-50), and \subref{fig:adv:d} Residual Network with 152 layers (ResNet-152). The class label of the adversarially perturbed ImageNet example and corresponding metrics are shown below images: perceptual adversarial similarity score (PASS) \cite{rozsa2016adversarial} between original and adversarial image pairs, followed by L$_2$ and L$_\infty$ norms of the adversarial perturbation. For better visualization, perturbations are scaled by a factor of 25 with gray indicating no change. Although all examples are formed by imperceptible perturbations for a human observer, adversarial examples originating from better performing models in \protect\subref{fig:adv:c} and \protect\subref{fig:adv:d} contain perturbations with higher magnitudes.}
\end{center}
\vspace*{0.1in}
\end{figure}

Szegedy et al.~\cite{szegedy2013intriguing} also analyzed the cross-model generalization of adversarial examples and concluded that ``a relatively large fraction of examples will be misclassified by networks trained from scratch with different hyper-parameters (number of layers, regularization  or initial weights)''. They quantified \textit{adversarial portability} -- the ability of adversarial examples generated on one model to fool another -- with respect to a particular non-convolutional model on the MNIST dataset \cite{lecun1998mnist}. Although both Szegedy et al.~\cite{szegedy2013intriguing} and Goodfellow et al.~\cite{goodfellow2014explaining} showed examples of adversarial images on ImageNet, they only performed quantitative evaluation on the MNIST dataset.

Since deep neural networks are vulnerable to attacks by applying small adversarial perturbations to their inputs, adversarial examples pose a potential security threat for the application of those machine learning models. The cross-model portability of adversarial examples indicate a more severe problem, namely, vulnerable machine learning systems can be attacked by causing misclassifications without having access to the underlying model.

In this paper, we study different types of adversarial examples generated on various learning models, and analyze adversarial portability across modern deep convolutional neural networks. The major contributions of this paper are:

\begin{enumerate}

\item We generate adversarial examples on the ImageNet dataset \cite{deng2009imagenet} by using three different low-cost techniques. We evaluate these techniques by analyzing the produced adversarial images with respect to various metrics.

\item We evaluate the robustness of eight deep neural networks -- including state-of-the-art models of recent years -- to adversarial examples by analyzing the metrics of the generated examples.

\item We experimentally evaluate the portability of adversarial examples across eight deep neural networks. We show that adversarial images are mostly portable across similar network topologies.

\item We find that best performing models are more difficult to be fooled, i.e., networks with higher accuracies require greater perturbations to form adversarial examples.

\end{enumerate}

\section{Related Work}

Adversarial examples in deep neural networks were discovered by Szegedy et al.~\cite{szegedy2013intriguing}. The authors demonstrated the first method to reliably find these perturbations by a box-constrained optimization technique (L-BFGS) that relies on internal network state. However, due to the computationally expensive L-BFGS optimization, this method is impractical. Furthermore, Szegedy et al. performed experiments on a few networks and datasets and concluded that a relatively large fraction of the adversarial examples were misclassified by different networks trained with varying hyperparameters or on disjoint training sets.

Goodfellow et al.~\cite{goodfellow2014explaining} presented the simpler and computationally cheaper fast gradient sign (FGS) algorithm for adversarial generation, and also demonstrated that machine learning models can benefit from these perturbed inputs. FGS uses the sign of the gradient of loss with respect to the input image to form adversarial perturbations. FGS is computationally efficient as the gradient can be effectively calculated using backpropagation, and the generated perturbations cause unexpected classification errors in various machine learning models. Based upon FGS and by simply using the raw gradient of loss, Rozsa et al.~\cite{rozsa2016adversarial} formalized the fast gradient value (FGV) method and introduced the hot/cold (HC) approach, which is capable of efficiently producing multiple adversarial examples for each input.

Papernot et al.~\cite{papernot2016limitations} introduced a method to produce adversarial perturbations by leveraging mappings between inputs and outputs of neural networks. Their experiments on the MNIST dataset suggest that identifying those mappings is computationally expensive. Baluja et al.~\cite{baluja2015virtues} proposed an approach which generates perturbations, applies them to input samples, and then observes how models respond to these perturbed images. They applied small affine image transformations to form perturbations without utilizing the internal state of the networks. Also, in order to filter out radical perturbations and identify useful ones for retraining, Baluja et al. used peer networks as a control-mechanism. Although these types of \emph{guess and check} approaches are capable of finding adversarial perturbations, they can be prohibitively expensive.


Sabour et al.~\cite{sabour2016adversarial} demonstrated that adversarial images can be produced by manipulating internal representations to mimic those that are present in targeted images. Their approach also relies on the inefficient L-BFGS algorithm. Moosavi-Dezfooli et al.~\cite{moosavi2016deepfool} introduced a method, which can produce adversarial perturbations with small L$_2$ or L$_\infty$ norms, but their approach is also computationally expensive.

Several approaches have been published to increase the robustness of vulnerable machine learning models to adversarial examples \cite{goodfellow2014explaining, luo2015foveation, papernot2015distillation, rozsa2016adversarial, miyato2016virtual, zheng2016improving}. These approaches can also lead to further improved overall performances as well.

Researchers also try to assess the severity of the security threat imposed by adversarial examples in the physical world \cite{kurakin2016adversarial} or even suggest attack strategies that rely on the transferability of such examples \cite{papernot2016transferability, papernot2016practical}. Specifically, Papernot et al.~\cite{papernot2016transferability, papernot2016practical} suggest a black-box attack by crafting adversarial samples on one model and apply those on the targeted oracle to cause misclassification. The authors report that such attacks can achieve high success rates with samples formed by adversarial perturbations with radically increased magnitudes. Since the ``informal'' definition of adversarial examples requires small perturbations to be applied on original inputs, we note that FGS examples with visually perceptible modifications cannot be considered adversarial in nature. Furthermore, recent advances indicate that those noisy samples would probably be classified as \textit{unknown} by open set deep networks \cite{bendale2016towards}, and -- depending on the dataset and context -- that might eliminate the threat posed by the introduced attack.

In this paper, we generate various types of adversarial examples formed by perturbations with minimal magnitudes that cause misclassifications, and we show on ImageNet that adversarial portability across deep convolutional neural networks is relatively low compared to prior work focusing on the MNIST dataset.

\section{Experiments}

Szegedy et al.~\cite{szegedy2013intriguing} demonstrated that the same adversarial images are often misclassified by a variety of learning models with different architectures or trained on varying training data. In order to have a better understanding of the adversarial portability problem for deep convolutional neural networks and to assess the adversarial robustness of those networks, we perform large-scale experiments with different types of adversarial images on various machine learning models. In this section, we introduce the models and the dataset that we use, and briefly describe our experiments.

\subsection{Models and Dataset}
\label{sec:models_dataset}

We test adversarial robustness and portability on the ILSVRC-2014 challenge dataset, which contains 1000 object categories from the large-scale hierarchical ImageNet \cite{deng2009imagenet} database with approximately 1.28M training images, 50K validation images, and 100K test images (test images are publicly not available). Performance of machine learning models is evaluated by top-1 and top-5 accuracies.

The deep neural networks that we use in our experiments are all publicly available: BVLC-AlexNet is a Berkeley-trained version of the model introduced by Krizhevsky et al.~\cite{krizhevsky2012imagenet}, VGG-16 and VGG-19 are 16-layer and 19-layer networks of Simonyan et al.~\cite{Simonyan15}, BVLC-GoogLeNet is the Berkeley-trained version of the network designed by Szegedy et al.~\cite{szegedy2015going}, Princeton-GoogLeNet is the GPU implementation of GoogLeNet by the Princeton Vision Group and, finally, ResNet-50, ResNet-101, and ResNet-152 are the 50-, 101-, and 152-layer networks of He et al.~\cite{he2015deep}, respectively.

To be able to test portability, we need all models to operate on images having the same dimensions. Since BVLC-AlexNet works with 227\,$\times$\,227 pixel images, while the others use 224\,$\times$\,224, we fine-tuned the model on the training set with the smaller crop size (a single epoch with the provided hyperparameters and a constant learning rate of $10^{-4}$). In the rest of the paper, we refer to this fine-tuned model as BVLC-AlexNet*.

The performance of the eight models on the ImageNet validation set is listed in Tab.~\ref{tab:errors}.
We obtained these error rates by using a single center crop from each of the 256\,$\times$\,256 scaled training images. Note that for some models, better performance can be achieved by using 10-crop error (averaging softmax scores of 10 224\,$\times$\,224 pixel crops) and/or by scaling images with shorter side to 256 pixels.

For each of the 1000 classes, we selected 10 images from the training set that were correctly classified by all eight models. Therefore, the dataset that we use for our experiments on adversarial images contains 10K images.

\begin{table}
\small
\centering\vspace*{1.5ex}
\renewcommand{\arraystretch}{1.05}
\begin{tabular}{@{\quad}c@{\quad}|@{\quad}c@{\quad}|@{\quad}c@{\quad}|@{\quad}c@{\quad}}  \small

ID & MODEL  				&	TOP-1 &	TOP-5	\vphantom{\large$|^x$} \\ \hline \hline
M1 & BVLC-AlexNet*		&	$43.230\%$		&	$20.012\%$  	\vphantom{\large$|^x$} \\ \hline
M2 & VGG-16				&	$31.642\%$		&   $11.556\%$ 	\vphantom{\large$|^x$} \\ \hline
M3 & VGG-19				&	$31.516\%$		&   $11.558\%$ 	\vphantom{\large$|^x$} \\ \hline
M4 & Princeton-GoogLeNet~	&	$32.934\%$ 		&   $12.104\%$  	\vphantom{\large$|^x$}	\\ \hline
M5 & BVLC-GoogLeNet 		&	$31.070\%$ 		&   $10.856\%$ 	\vphantom{\large$|^x)$} 	\\ \hline
M6 & ResNet-50	 		&	$27.124\%$ 		&   $8.864\%$ 	\vphantom{\large$|^x)$}	\\ \hline
M7 & ResNet-101	 		&	$25.656\%$ 		&   $8.054\%$ 	\vphantom{\large$|^x)$}	\\ \hline
M8 & ResNet-152 			&	$25.090\%$ 		&   $7.798\%$ 	\vphantom{\large$|^x)$}	\\

\end{tabular}
\cap{tab:errors}{Error Rates}{This table lists the top-1 and top-5 classification error rates of the investigated models on the ImageNet validation dataset. For consistency, we report error rates on 224\,$\times$\,224 pixel center crops from 256\,$\times$\,256 scaled images.}
\end{table}

\subsection{Adversarial Generation}

In order to evaluate the adversarial robustness of the selected models and quantify adversarial portability across them, we use three adversarial example generation methods. First, the fast gradient sign (FGS) method introduced by Goodfellow et al.~\cite{goodfellow2014explaining}. Second, the fast gradient value (FGV) approach formalized in \cite{rozsa2016adversarial}, which is built upon FGS. While FGS takes steps in the direction defined by the sign of the gradient of loss with respect to the image in order to reduce the score of the correct class and cause mislabeling, FGV uses a scaled version of the raw gradient of loss and produces notably different adversarial perturbations than FGS. Third, the hot/cold (HC) approach introduced by Rozsa et al.~\cite{rozsa2016adversarial}, which is capable of generating multiple diverse adversarial samples from an input by not only reducing the score of the original correct class -- denoted as the \emph{cold class} -- but in parallel by increasing the score of a selected \emph{hot class}. Specifically, we use the hot/cold approach with only the most similar class with respect to the classification score as \emph{hot} (HC1) and, hence, we generate only one adversarial example for each input with this technique. We have selected these three adversarial generation methods because they are computationally efficient and also able to produce diverse samples.

We commence our experiments generating adversarial images on our dataset of 10K images. We use the three aforementioned adversarial generation methods (FGS, FGV, and HC1) with the previously listed eight deep neural networks and collect various metrics on the produced samples to compare the adversarial generation methods. As pointed out by Sabour el al.~\cite{sabour2016adversarial}, L$_2$ and L$_\infty$ norms of adversarial perturbations are not matched well to human perception. Since these measures are extremely sensitive to even small geometric distortions and do not map well to psychophysical notions of similarity, we also use the perceptual adversarial similarity score (PASS) \cite{rozsa2016adversarial} to better quantify ``adversarial'' in terms of human perception. Additionally, we measure the adversarial success rate, i.e., the relative number of images for which an adversarial example can be generated. Adversarial image generation can fail in two ways. First, the adversarial direction (e.g., the gradient of loss with respect to the input image used by FGS and FGV methods) can be exactly zero for all pixels. Second, any arbitrarily large step into the adversarial direction -- limited by the discrete pixel values in range $[0,255]$ -- does not change the original label.

\begin{figure*}[t!]
\begin{center}
\centering\subfloat[][\label{fig:overall_met:a}\centering \textit{PASS} and success rates for various types of adversarial images]{\includegraphics[width=1.56\columnwidth]{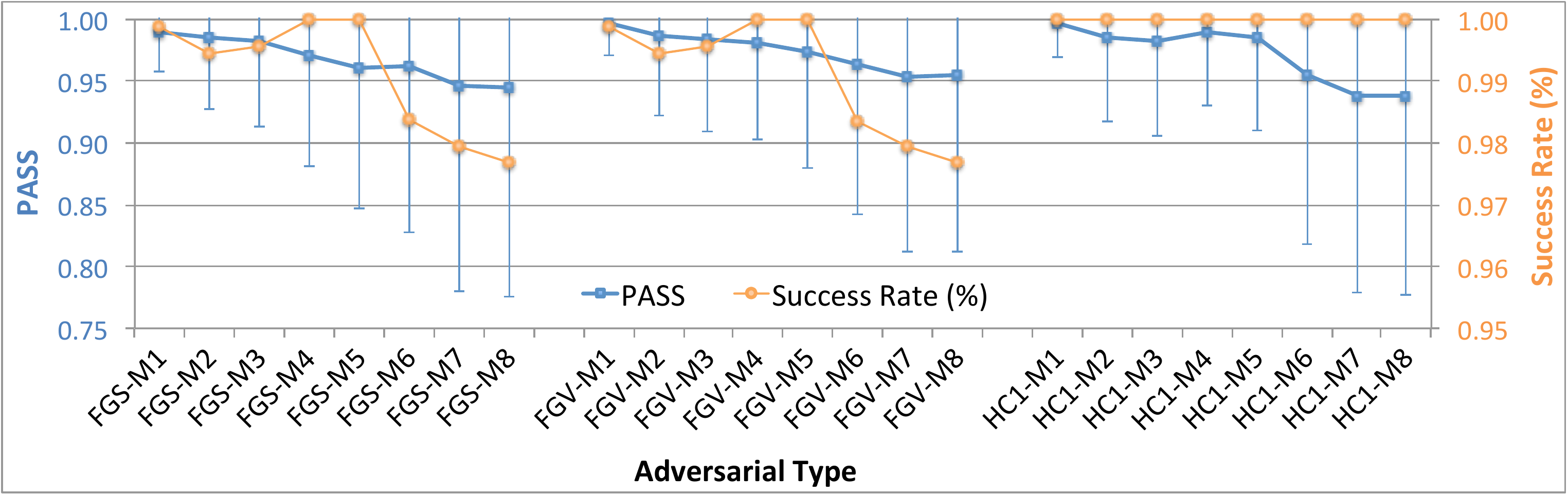}} \\

\centering\subfloat[][\label{fig:overall_met:b}\centering $L_\infty$ and $L_2$ norms for various types of adversarial images]{\includegraphics[width=1.56\columnwidth]{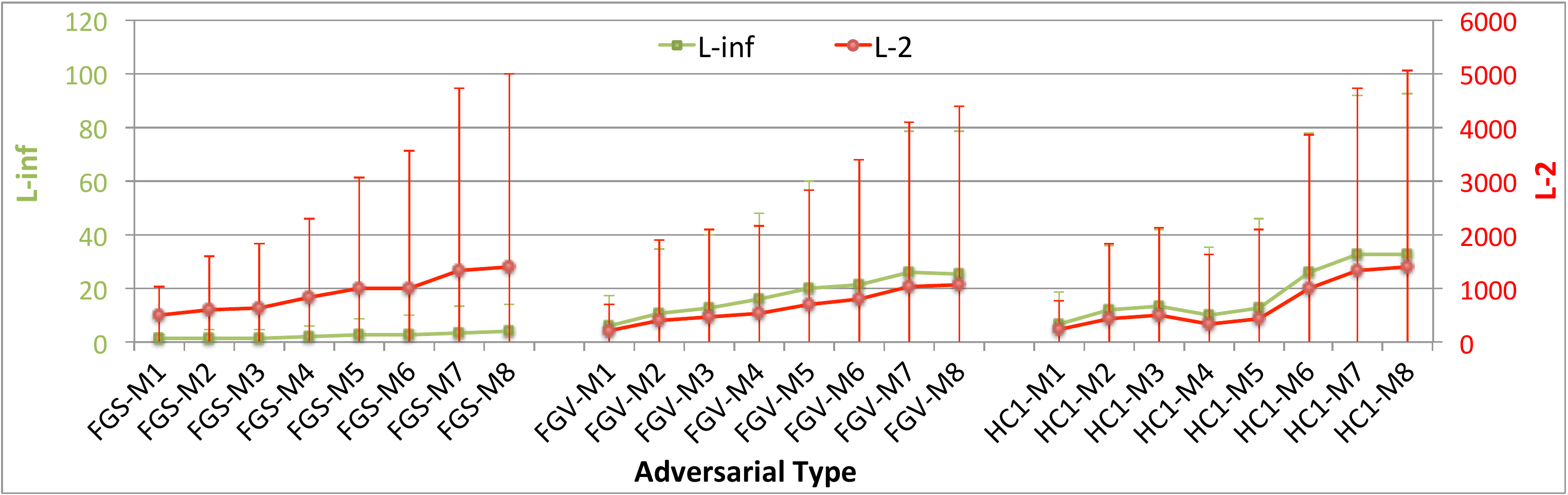}}

\cap{fig:overall_met}{Metrics for Adversarial Images}{This figure shows various metrics for adversarial images generated by fast gradient sign (FGS), fast gradient value (FGV), and hot/cold (HC1) approaches on the networks listed in Tab.~\ref{tab:errors} using 10K images of the ImageNet training set. In \subref{fig:overall_met:a} the mean and standard deviations of PASS scores and the adversarial success rates are displayed, while in \subref{fig:overall_met:b} the means and standard deviations of $L_2$ and $L_\infty$ norms are presented.}
\end{center}
\end{figure*}

As we can see in Fig.~\ref{fig:overall_met}\subref{fig:overall_met:a}, all three adversarial generation methods maintain high success rates in terms of producing adversarial images for their inputs on various deep neural networks. Particularly, HC1 reaches approximately 100\,\% success rate on each model. Based on the metrics shown in Fig.~\ref{fig:overall_met}, we can observe that FGV and HC1 methods produce adversarial perturbations with significantly higher $L_\infty$ norms, while the formed adversarial images still maintain comparable or even higher PASS scores than others generated by FGS.


\subsection{Adversarial Robustness}

By investigating the collected metrics of adversarial images generated by various methods on given deep neural networks, we can compare the robustness of those networks to adversarial examples. Specifically, increased L$_2$ and L$_\infty$ norms of adversarial perturbations paired with decreased PASS scores indicate higher robustness. In other words, models that require samples formed by adversarial perturbations with higher magnitudes in terms of L$_2$ and L$_\infty$ norms to cause misclassifications are more robust to adversarial examples.

As shown by the collected metrics in Fig.~\ref{fig:overall_met}\subref{fig:overall_met:b}, L$_2$ and L$_\infty$ norms of adversarial perturbations generally increase as we generate adversarial examples on better performing machine learning models. We can see in in Fig.~\ref{fig:overall_met}\subref{fig:overall_met:a} that these stronger perturbations also result in decreased PASS scores. Based on these results, we can conclude that better performing models lead to increased robustness to adversarial examples as well. To better illustrate this phenomenon, we show FGV adversarial examples with corresponding perturbations in Fig.~\ref{fig:adv} crafted from the same input image on four different models. Although these samples contain imperceptible modifications to the human eye, we can observe that a significantly stronger perturbation is required to cause misclassification on the current state-of-the-art Residual Network, denoted as ResNet-152 in Fig.~\ref{fig:adv}\subref{fig:adv:d}, compared to VGG-16 shown in Fig.~\ref{fig:adv}\subref{fig:adv:a}. Finally, we can state that not only the magnitudes and structures of these perturbations vary, but they can lead to different misclassifications. Depending on the model, the same adversarial image generation approach turns the \textit{tiger shark} into a \textit{white shark} or a \textit{hammerhead}.

\subsection{Adversarial Portability}

To evaluate the transferability of adversarial examples across deep convolutional neural networks, we use the different types of adversarial images -- formed by using either the fast gradient sign (FGS) method, the fast gradient value (FGV) method, or the hot/cold approach with the most similar class in terms of classification score (HC1) -- that we generated on the dataset described in Sec. \ref{sec:models_dataset}. Since FGS, FGV, and HC1 methods all rely on internal network state, the perturbations that form adversarial images on various networks can be significantly different -- as demonstrated by the FGV examples shown in Fig.~\ref{fig:adv} that were generated on a \textit{tiger shark}.

\begin{figure*}[t!]
\begin{center}
\centering\includegraphics[width=.99\textwidth]{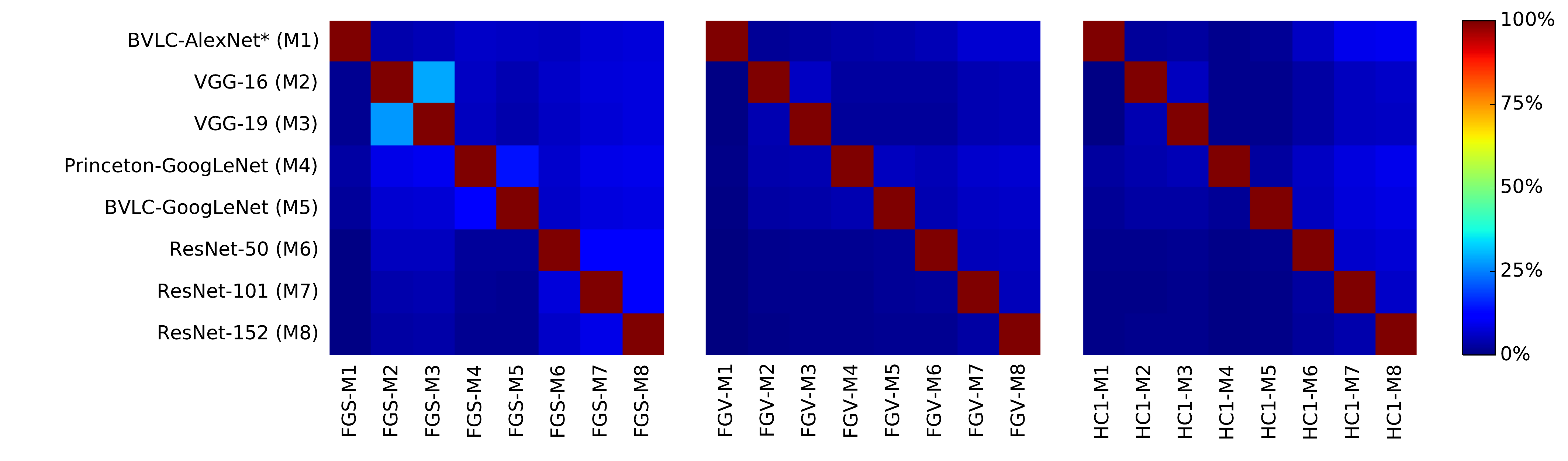}

\cap{fig:confusion}{Cross-Model Portability of Adversarial Examples}{This figure shows the percentage of adversarial examples generated on one network that are adversarial to other networks. Adversarial examples of types FGS, FGV, and HC1 are generated on the learning models as listed in Tab.~\ref{tab:errors} (shown on the horizontal axis), and tested on the networks displayed in the vertical axis. The diagonals display the self-portability of adversarial examples, which is 100\,\% in each case.}
\end{center}
\end{figure*}

To quantify adversarial portability, we test the constructed adversarial examples across the eight deep convolutional neural networks listed in Tab.~\ref{tab:errors}, and calculate the proportion of transferable samples. The graphically summarized results are presented in Fig.~\ref{fig:confusion}.
Apparently, adversarial images originating from similar models or models sharing the same network architecture are generally more portable across corresponding machine learning models. For example, 29.00\,\% of FGS adversarial images generated on the VGG-16 network (denoted as FGS-M2 in Fig.~\ref{fig:confusion}) remain adversarial on the VGG-19 model, and 27.59\,\% of VGG-19 samples (FGS-M3) are transferable to VGG-16. Similar patterns can be observed between the BVLC-GoogLeNet and the Princeton-GoogLeNet models: 14.23\,\% of the BVLC-GoogLeNet FGS adversarial images (FGS-M5) are transferable to Princeton-GoogLeNet and 12.88\,\% remain portable in the opposite direction (FGS-M4). Although the portability rates between Residual Networks are lower, they exhibit the same behavior: 12.13\,\% of FGS samples originating from ResNet-101 (FGS-M7) cause misclassifications on ResNet-50, while 7.97\,\% of adversarial samples are portable backwards (FGS-M6).

Fig.~\ref{fig:confusion} displays that the tested models are more robust to FGV and HC1 samples than to FGS samples. We believe that the higher portability of FGS samples is due to the application of the sign in the fast gradient sign method, which provides equal importance to pixels regardless their actual raw value in the gradient of loss. Specifically, FGS applies unnecessarily large modifications on pixels that do not play an important role in reducing the loss, and these redundant perturbations make FGS examples more transferable. Therefore, the portability rates for FGV and HC1 adversarial images are significantly lower than for FGS samples. The most transferable adversarial images among FGV and HC1 are HC1 samples of the ResNet-152 model (denoted as HC1-M8 in Fig.~\ref{fig:confusion}) tested on BVLC-AlexNet* network with 9.99\,\% portability rate. In other words, HC1 adversarial samples generated on the best performing network cause the most classification errors on the worst performing model. On the other hand, adversarial examples generated on BVLC-AlexNet* are nearly never portable to other networks.

As shown in Fig.~\ref{fig:confusion}, adversarial examples generated on the best performing networks, i.e., the Residual Networks, are generally more portable to others. Since those networks are the most robust among the tested models, their adversarial examples contain the strongest perturbations and, therefore, we assume that those samples become more transferable.

In summary, the results suggest that FGS examples are more transferable than FGV or HC1 samples, and adversarial images are mainly portable across similar networks, e.g., VGG models (VGG-16, VGG-19), GoogleNets (BVLC-GoogLeNet, Princeton-GoogLeNet) or Residual Networks (ResNet-50, ResNet-101, ResNet-152).
However, the tranferability rates that we obtained on the ImageNet dataset are considerably lower than Szegedy et al.~\cite{szegedy2013intriguing} observed when they analyzed cross-model generalization of adversarial examples on the MNIST dataset.

\section{Discussion}

In this paper, we have generated adversarial examples on eight deep convolutional neural networks, including state-of-the-art models of recent years, with three different adversarial generation methods. We have evaluated these methods with respect to their success rates, and quantitatively compared various types of adversarial images. We have found that all three methods -- fast gradient sign (FGS), fast gradient value (FGV), and hot/cold (HC) -- can efficiently produce adversarial samples on all tested models.

By analyzing the collected metrics -- L$_2$ and L$_\infty$ norms of adversarial perturbations, and perceptual adversarial similarity score (PASS) of original and adversarial image pairs -- on the tested models, we have observed that better performing networks are more robust to adversarial perturbations. We believe that more accurate deep convolutional neural networks learn feature mappings for the given classification task that make classes more separable and, therefore, these models generalize better, leading to both higher accuracies and improved robustness.

We have performed large-scale experiments on the ImageNet dataset to investigate the portability of various types of adversarial images across multiple deep convolutional neural networks including the fine-tuned version of BVLC-AlexNet, the publicly available versions of VGG-16 and VGG-19, Princeton-GoogLeNet, BVLC-GoogLeNet, and three versions of Residual Network models. As our results suggest, adversarial images are more transferable between networks sharing the same or similar network architectures, and adversarial samples originating from better performing models are more transferable to more vulnerable networks due to their stronger adversarial perturbations. Furthermore, our experiments have highlighted that adversarial images generated by FGS are more portable than others produced by FGV or HC.

In our cross-model transferability experiments, we have used adversarial examples formed by perturbations with minimal magnitudes that cause misclassifications. We note that sometimes these perturbations are highly perceptible and, therefore, the crafted samples cannot be considered adversarial. While those noisy samples are not adversarial in nature, they still have an effect on our measured portability rates as, due to their stronger perturbations, they are certainly more portable across networks. The perceptual adversarial similarity score (PASS) seems to be the straightforward measure to differentiate adversarial from noisy samples, i.e., to quantitatively define adversarialness by measuring the similarity/distinguishability of original and perturbed image pairs. Specifying an applicable threshold for PASS scores to define \textit{adversarial} is beyond the scope of this paper.


There are already applied mechanisms to inadvertently mitigate the transferability problem of adversarial examples. To achieve better performances on various visual recognition tasks, machine learning systems often use multiple crops for classification. Namely, several crops are extracted from the input and after classifying them independently, the model makes the final classification by fusing the results obtained from the crops. Luo et al.~\cite{luo2015foveation} proposed a mechanism to alleviate the recognition errors caused by adversarial images, which is basically based upon cropping. Their foveation-based technique uses only a sub-region of the image during classification. The authors demonstrated that the negative effect of foveated perturbations to classification scores can be significantly reduced compared to entire perturbations, suggesting that foveation approaches can improve the robustness of neural networks to adversarial examples. Another popular approach to obtain better classification performances is the application of multiple models by using an ensemble of networks. For instance, He et al.~\cite{he2015deep} applied an ensemble of three Residual Networks. As we demonstrated, the proportion of portable adversarial images among Residual Networks is relatively low. Therefore, the application of ensembles can further improve the robustness to adversarial examples.


In summary, considering our experimental results and the aforementioned techniques applied in real-world applications, we conclude that the security threat posed by adversarial portability is moderate, at most, but this area is still important for applications and future research.

\section*{Acknowledgment}

This research is based upon work funded in part by NSF IIS-1320956 and in part by the Office of the Director of National Intelligence (ODNI), Intelligence Advanced Research Projects Activity (IARPA), via IARPA R\&D Contract No. 2014-14071600012. The views and conclusions contained herein are those of the authors and should not be interpreted as necessarily representing the official policies or endorsements, either expressed or implied, of the ODNI, IARPA, or the U.S. Government. The U.S. Government is authorized to reproduce and distribute reprints for Governmental purposes notwithstanding any copyright annotation thereon.

\bibliographystyle{IEEEtran}
\bibliography{lots}

\begin{thebibliography}{10}
\providecommand{\url}[1]{#1}
\csname url@samestyle\endcsname
\providecommand{\newblock}{\relax}
\providecommand{\bibinfo}[2]{#2}
\providecommand{\BIBentrySTDinterwordspacing}{\spaceskip=0pt\relax}
\providecommand{\BIBentryALTinterwordstretchfactor}{4}
\providecommand{\BIBentryALTinterwordspacing}{\spaceskip=\fontdimen2\font plus
\BIBentryALTinterwordstretchfactor\fontdimen3\font minus
  \fontdimen4\font\relax}
\providecommand{\BIBforeignlanguage}[2]{{%
\expandafter\ifx\csname l@#1\endcsname\relax
\typeout{** WARNING: IEEEtran.bst: No hyphenation pattern has been}%
\typeout{** loaded for the language `#1'. Using the pattern for}%
\typeout{** the default language instead.}%
\else
\language=\csname l@#1\endcsname
\fi
#2}}
\providecommand{\BIBdecl}{\relax}
\BIBdecl

\bibitem{szegedy2015going}
C.~Szegedy, W.~Liu, Y.~Jia, P.~Sermanet, S.~Reed, D.~Anguelov, D.~Erhan,
  V.~Vanhoucke, and A.~Rabinovich, ``Going deeper with convolutions,'' in
  \emph{IEEE Conference on Computer Vision and Pattern Recognition (CVPR)},
  2015, pp. 1--9.

\bibitem{he2015delving}
K.~He, X.~Zhang, S.~Ren, and J.~Sun, ``Delving deep into rectifiers: Surpassing
  human-level performance on {ImageNet} classification,'' in \emph{IEEE
  International Conference on Computer Vision (ICCV)}, 2015, pp. 1026--1034.

\bibitem{chen2016unconstrained}
J.-C. Chen, V.~M. Patel, and R.~Chellappa, ``Unconstrained face verification
  using deep {CNN} features,'' in \emph{IEEE Winter Conference on Applications
  of Computer Vision (WACV)}, 2016.

\bibitem{he2015deep}
K.~He, X.~Zhang, S.~Ren, and J.~Sun, ``Deep residual learning for image
  recognition,'' in \emph{IEEE Conference on Computer Vision and Pattern
  Recognition (CVPR)}, 2016.

\bibitem{szegedy2013intriguing}
C.~J. Szegedy, W.~Zaremba, I.~Sutskever, J.~Bruna, D.~Erhan, I.~Goodfellow, and
  R.~Fergus, ``Intriguing properties of neural networks,'' in
  \emph{International Conference on Learning Representation (ICLR)}, 2014.

\bibitem{rozsa2016adversarial}
A.~Rozsa, E.~M. Rudd, and T.~E. Boult, ``Adversarial diversity and hard
  positive generation,'' in \emph{IEEE Conference on Computer Vision and
  Pattern Recognition (CVPR) Workshops}, 2016.

\bibitem{Simonyan15}
K.~Simonyan and A.~Zisserman, ``Very deep convolutional networks for
  large-scale image recognition,'' in \emph{Proceedings of the International
  Conference on Learning Representations (ICLR)}, 2015.

\bibitem{lecun1998mnist}
Y.~LeCun, C.~Cortes, and C.~J. Burges, ``The mnist database of handwritten
  digits,'' 1998.

\bibitem{goodfellow2014explaining}
I.~J. Goodfellow, J.~Shlens, and C.~Szegedy, ``Explaining and harnessing
  adversarial examples,'' in \emph{International Conference on Learning
  Representation (ICLR)}, 2015.

\bibitem{deng2009imagenet}
J.~Deng, W.~Dong, R.~Socher, L.-J. Li, K.~Li, and L.~Fei-Fei, ``{ImageNet}: A
  large-scale hierarchical image database,'' in \emph{IEEE Conference on
  Computer Vision and Pattern Recognition (CVPR)}, 2009, pp. 248--255.

\bibitem{papernot2016limitations}
N.~Papernot, P.~McDaniel, S.~Jha, M.~Fredrikson, Z.~B. Celik, and A.~Swami,
  ``The limitations of deep learning in adversarial settings,'' in \emph{IEEE
  European Symposium on Security and Privacy (Euro S\&P)}, 2016, arXiv preprint
  arXiv:1511.07528.

\bibitem{baluja2015virtues}
S.~Baluja, M.~Covell, and R.~Sukthankar, ``The virtues of peer pressure: A
  simple method for discovering high-value mistakes,'' in \emph{Computer
  Analysis of Images and Patterns (CAIP)}.\hskip 1em plus 0.5em minus
  0.4em\relax Springer, 2015, pp. 96--108.

\bibitem{sabour2016adversarial}
S.~Sabour, Y.~Cao, F.~Faghri, and D.~J. Fleet, ``Adversarial manipulation of
  deep representations,'' in \emph{International Conference on Learning
  Representations (ICLR)}, 2016.

\bibitem{moosavi2016deepfool}
S.-M. Moosavi-Dezfooli, A.~Fawzi, and P.~Frossard, ``{DeepFool}: a simple and
  accurate method to fool deep neural networks,'' in \emph{IEEE Conference on
  Computer Vision and Pattern Recognition (CVPR)}, 2016.

\bibitem{luo2015foveation}
Y.~Luo, X.~Boix, G.~Roig, T.~Poggio, and Q.~Zhao, ``Foveation-based mechanisms
  alleviate adversarial examples,'' 2015, under review.

\bibitem{papernot2015distillation}
N.~Papernot, P.~McDaniel, X.~Wu, S.~Jha, and A.~Swami, ``Distillation as a
  defense to adversarial perturbations against deep neural networks,'' in
  \emph{IEEE Symposium on Security and Privacy (SP)}, 2015.

\bibitem{miyato2016virtual}
T.~Miyato, A.~M. Dai, and I.~Goodfellow, ``Virtual adversarial training for
  semi-supervised text classification,'' 2016, under review.

\bibitem{zheng2016improving}
S.~Zheng, Y.~Song, T.~Leung, and I.~Goodfellow, ``Improving the robustness of
  deep neural networks via stability training,'' 2016, under review.

\bibitem{kurakin2016adversarial}
A.~Kurakin, I.~Goodfellow, and S.~Bengio, ``Adversarial examples in the
  physical world,'' Google, Tech. Rep., 2016.

\bibitem{papernot2016transferability}
N.~Papernot, P.~McDaniel, and I.~J. Goodfellow, ``Transferability in machine
  learning: from phenomena to black-box attacks using adversarial samples,''
  2016, under review.

\bibitem{papernot2016practical}
N.~Papernot, P.~McDaniel, I.~J. Goodfellow, S.~Jha, Z.~Berkay~Celik, and
  A.~Swami, ``Practical black-box attacks against deep learning systems using
  adversarial examples,'' 2016, under review.

\bibitem{bendale2016towards}
A.~Bendale and T.~Boult, ``Towards open set deep networks,'' in \emph{IEEE
  Conference on Computer Vision and Pattern Recognition (CVPR)}, 2016.

\bibitem{krizhevsky2012imagenet}
A.~Krizhevsky, I.~Sutskever, and G.~E. Hinton, ``{ImageNet} classification with
  deep convolutional neural networks,'' in \emph{Advances in Neural Information
  Processing Systems (NIPS)}, 2012, pp. 1097--1105.

\end{thebibliography}

\end{document}